\documentclass{article}
\usepackage{orcidlink}
\usepackage{amsmath,amssymb,amsfonts}
\usepackage[margin=4cm]{geometry}
\usepackage{algpseudocode}
\usepackage{algorithm}
\usepackage{algorithmicx}
\usepackage{arxiv}
\usepackage[utf8]{inputenc} % allow utf-8 input
\usepackage[T1]{fontenc}    % use 8-bit T1 fonts
\usepackage{hyperref}       % hyperlinks
\usepackage{url}            % simple URL typesetting
\usepackage{graphicx}
\usepackage{plantuml}
\usepackage{multirow}
\usepackage{booktabs}       % professional-quality tables
\usepackage{flafter}		% for formatting tables correctly (tables stuck at top / not under section  issue)
\usepackage{longtable}
\usepackage{acronym}
\usepackage[acronym]{glossaries}
\usepackage{diagbox}

\newacronym{SR GAN}{SR GAN}{Super Resolution Generative Adversarial Networks}
\newacronym{LR}{LR}{Low Resolution}
\newacronym{OCR}{OCR}{Optical Character Recognition}
\newacronym{SR}{SR}{Super Resolution}
\newacronym{PReLU}{PReLU}{Parametric Rectifier Linear Unit}
\newacronym{BN}{BN}{Bacth Normalization}

\graphicspath{ {./images/} }
\title{ A comparative analysis of SRGAN models }

\author{ 
    \href{https://orcid.org/0009-0002-4952-7762}
    {\includegraphics[scale=0.06]{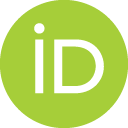}\hspace{1mm}Fatemeh Rezapoor Nikroo$^*$}\\
	Synechron Data Practice (Montreal, Canada)\\
    \texttt{Fatemeh.Nikroo@synechron.com}\\
    \And
    \href{https://orcid.org/0009-0004-1722-0306}
    {\includegraphics[scale=0.06]{orcid}\hspace{1mm}Ajinkya Deshmukh$^*$}\\
	Synechron Data Practice (Pune, India)\\
    \texttt{Ajinkya.Deshmukh@synechron.com}\\
    \And
    \href{https://orcid.org/0009-0007-1461-1231}
    {\includegraphics[scale=0.06]{orcid}\hspace{1mm}Kaarthik Senthil Kumar$^*$}\\
	Synechron Data Practice (New York, USA)\\
    \texttt{Kaarthik.Kumar@synechron.com}
    \And  
    \href{https://orcid.org/0009-0002-9433-5903}
    {\includegraphics[scale=0.06]{orcid}\hspace{1mm} Cleo Norris$^*$}\\
	Synechron Data Practice (Montreal, Canada)\\
    \texttt{Cleo.Norris@synechron.com}
    \And
    \href{https://orcid.org/0000-0002-9064-3362}
    {\includegraphics[scale=0.06]{orcid}\hspace{1mm}Anantha Sharma}\\
	Synechron Data Practice (Charlotte, USA)\\
    \texttt{Anantha.Sharma@synechron.com}
    \And
    \href{https://orcid.org/0000-0002-9067-5663}
    {\includegraphics[scale=0.06]{orcid}\hspace{1mm}Adrian Tam}\\
	Synechron Data Practice (New York, USA)\\
    \texttt{Adrian.Tam@synechron.com}
    \And
    \href{https://orcid.org/0009-0006-4508-1950}
    {\includegraphics[scale=0.06]{orcid}\hspace{1mm}Aditya Dangi}\\
	Synechron Data Practice (London, UK)\\
    \texttt{Aditya.Dangi@synechron.com}
}
     \renewcommand{\headeright}{}
 \renewcommand{\undertitle}{Synechron Data Practice}

\begin{document}

\maketitle

\def\thefootnote{*}\footnotetext{These authors contributed equally to this work.}

\begin{abstract}
%\footnote{* all authors contributed equally to this work.}
In this study, we evaluate the performance of multiple state-of-the-art SRGAN (Super Resolution Generative Adversarial Network) models, ESRGAN, Real-ESRGAN and EDSR, on a benchmark dataset of real-world images which undergo degradation using a pipeline. Our results show that some models seem to significantly increase the resolution of the input images while preserving their visual quality, this is assessed using Tesseract OCR engine. We observe that EDSR-BASE model from huggingface outperforms the remaining candidate models in terms of both quantitative metrics and subjective visual quality assessments with least compute overhead. Specifically, EDSR generates images with higher peak signal-to-noise ratio (PSNR) and structural similarity index (SSIM) values and are seen to return high quality OCR results with Tesseract OCR engine. These findings suggest that EDSR is a robust and effective approach for single-image super-resolution and may be particularly well-suited for applications where high-quality visual fidelity is critical and optimized compute. 
\end{abstract}

\section{Introduction}
Super resolving text before applying text recognition methods can increase the likelihood of recognizability and needs to be addressed \cite{tajik2021ideal}. Image super-resolution (SR), particularly the single image super-resolution problem, has received a lot of attention in the computer vision research community \cite{lim2017enhanced}. The advent of deep learning convolutional neural networks (CNNs) has revolutionized the SR problem and introduced remarkable improvements in this field \cite{lim2017enhanced}. 
 Studies have shown that photo-realism is usually attained by adversarial training using GANs \cite{NIPS2014_5ca3e9b1}. We evaluated the EDSR model (which employs deep CNNs) and found that it performs comparably to GANs; GAN networks have the ability to generate new samples that look real and can be used in the context of image resolution \cite{goodfellow2014generative}, while the EDSR model aims at purely reconstructing images by capturing high-frequency details.  

In this work we compare the capability of four models: 

EDSR \cite{eugenesiowEDSR} 

EDSR-base (Enhanced Deep Super-Resolution) \cite{lim2017enhanced,eugenesiowEDSRBase}

ESRGAN (Enhanced Super-Resolution Generative Adversarial) \cite{wang2018esrgan} 

Real-ESRGAN (upgraded ESRGAN) \cite{wang2021realesrgan} 

The goal here is to accurately OCR (Optical Character Recognition) text from these images, using GANs to improve image quality and assessing improvements in OCR accuracy. We are using 2x, 3x and 4x upsampling factors to enhance the resolution of degraded text image before running OCR extraction and subsequent comparison. 
\section{Introduction to candidate GAN Models}
\subsection{ Background}
Candidate models incorporate the concept of GANs (Generative Adversarial Networks), a deep learning framework with two main components of a generator and discriminator. 

These two neural networks train simultaneously in an adversarial manner. The generator network is trained to produce synthetic meaningful data in line with the input data distribution, while the discriminator network acts as a binary classifier to distinguish real data from the generated one. 

Super-Resolution Generative Adversarial Network \cite{ledig2017photorealistic} is used for image super-resolution as an improvement over traditional single-image super-resolution techniques. 

In this section, we describe the GAN \cite{ledig2017photorealistic} algorithm and is outlined below.

\begin{algorithm}
\caption{Super-Resolution using GAN Algorithm}
\label{alg:example}
\begin{algorithmic}[1]
    \State \textbf{Input:low-resolution image} 
    \State \textbf{Output:High-resolution image} 
    \textbf{Step 1:} Initialize the GAN generator and discriminator networks\;
    \While{not converged}
        \State \textbf{Step 2:} Generate a high-resolution image using the generator\;
        \State \textbf{Step 3:} Train the discriminator using real and generated high-resolution images\;
        \State \textbf{Step 4:} Train the generator to generate high-resolution images to reduce the loss function\;
    \EndWhile
    
    \State \textbf{Step 6:} Obtain the high-resolution image using the trained generator\;

\end{algorithmic}
\end{algorithm}

The SR training algorithm in general starts with a 64x64 pixel low-resolution as input. 

\textbf{Step 1:} The SR generator and discriminator networks are initialized \cite{ledig2017photorealistic}. 

\textbf{Step 2:} The algorithm enters a while loop to train the two networks to achieve a more accurate model to recover the finer texture details of the single image. 

Here we focus on reducing loss rather than improving accuracy. The loss functions used are described later and are the main factors to retrain the generator and discriminator networks. In this regard, convergence of the algorithms to end the while loop is set to the stabilization of the generator and discriminator losses.

By following these steps, the SR algorithm transforms a low-resolution image into a super-resolution (SR) image with improved visual quality and enhanced details.

\subsection{Generator Network Architecture}
SR Generator network consists of multiple layers as below: 
\begin{itemize}
    \item Input Layer
    \item Convolutional Layer
    \item Residual Blocks
    \item Upsampling Layers
    \item Activation Functions
    \item Output Layer
\end{itemize}

The input is the low-resolution image which then goes to the first convolutional layer and produces a feature map. 

The next step is the activation function of \textbf{PReLU} (Parametric Rectifier Linear Unit), which can handle a wide range of complex inputs. The parametric aspect allows negative slope for negative values rather than considering them as zero. This way the network can learn from both positive and negative values. This enables better capturing of features.

The next step is the residual block, which consists of multiple layers. Each residual block has six layers in order as follows: 

\begin{itemize}
    \item Convolution 
    \item BN (Batch Normalization)
    \item PReLU 
    \item Convolution
    \item BN (Batch Normalization)
    \item Skip Connection (Elementwise Sum)
\end{itemize}

This architecture consists of 16 residual blocks stacked together with a skip connection technique to connect the activation layer to subsequent layers and skipping some layers in between. 
The skip connection is the elementwise sum of the input which comes from activation layer and the output of the block. 

Batch normalization is applied after convolution and before the activation layer. This enables us to use higher learning rates while allowing a wider range of initialization parameters \cite{ioffe2015batch}. 

After the input goes through the residual blocks, we have three more blocks. 
The first one consists of 

1) Convolution

2) BN 

3) Elementwise Sum 

The two other blocks both have 

1) Convolution

2) 2x PixelShuffler

3) PReLU

The \textbf{PixelShuffler} is the upscaling component which upscales the LR by two in each block and subsequently the LR image is upscaled by four before feeding it to the last convolutional layer. The output image is supposed to resemble the ground truth (as a high-resolution image).

\subsection{Discriminator Network Architecture}
The discriminator function is trained to distinguish super-resolution images from real images \cite{ledig2017photorealistic}. Input is either a high-resolution image from the generator network or an image from the training dataset. Likewise, the generator consists of a number of convolution layers to extract features. 

The first step is a convolution layer followed by a Leaky ReLU activation. 

The next step consists of eight blocks of three layers, each of which have 

1) convolution layer
2) BN
3) Leaky ReLU activation

To classify the input image we need to use a dense layer to flatten the multi-dimensional feature maps into a one-dimensional vector representation. This vector, followed by an activation layer, goes through another dense layer to convert the one-dimensional vector of size 1024 to one as the result. We applied the \textbf{Sigmoid} function to reduce input value to either zero or one, which reflects binary classes.

\subsection{ESRGAN}
Enhanced Super Resolution Generative Adversarial Network is a generalization which is able to better extract realistic and natural textures of LR images \cite{wang2018esrgan}. The algorithm follows similar steps as SRGAN with modifications in the network structure namely in the discriminator and generator \cite{wang2018esrgan} networks. 

The modifications in ESRGAN are as follows: 
\begin{itemize}
    \item BN components are omitted
    \item Residual Blocks are substituted with the Residual-in-Residual Dense Blocks (RRDB)
\end{itemize}
BN is removed to reach the stable training process and increase performance \cite{wang2018esrgan}. This comes from the fact that BN normalizes the features according to the mean and variance of the whole training dataset. Therefore, in cases with different underlying distributions of train and test data, this layer fails to predict accurately, since it only considers the statistics of the training dataset while predicting on the test dataset. According to this paper \cite{wang2018esrgan}, removing the BN layer is  key to reducing computational complexity and memory usage while making the model capable of high-quality reconstructions in a wide variety of images. Furthermore,  residual scaling and smaller initialization are considered in the proposed model to simplify the training process \cite{wang2018esrgan}. This study \cite{lim2017enhanced,zhang2018image,zhang2018residual} indicates that a higher number of layers and connections leads to better performance . The ESRGAN model introduced deep RRDB (residual-in-residual dense block) architecture with multi-level residual networks. 

\subsection{Generator Network Architecture-ESRGAN}
This uses RRDBs to stabilize training in deeper networks \cite{wang2018esrgan}. 

The layers are as follows: 
\begin{itemize}
    \item Input Layer
    \item Convolutional Layer
    \item RRDBs
    \item Upsampling
    \item  Output Layer
\end{itemize}
The LR image is fed as input into the model and is followed by a convolutional layer. It then enters the basic blocks which are RRDBs. 

In this paper \cite{wang2018esrgan}, 23 RRDBs are considered and can be modified based on the complexity of the problem. The last steps are upsampling  followed by  two consecutive convolutional layers. 

\textbf{RRDB} is a densely connected convolutional layer with residual connections to preserve low-level details. RRDBs are made of dense blocks, each of which has a convolution layer followed by a \textbf{Leaky ReLU} activation and one extra convolution at the end of the block. 

\subsection{Discriminator Network Architecture-ESRGAN}
The ESRGAN's discriminator is based on Relativistic GAN \cite{jolicoeurmartineau2018relativistic}, which estimates the probability of real data being more realistic than generated fake data. 

In Relativistic GAN compared to standard GAN, the output of the discriminator indicates the relative probability of it being real or fake data rather than using an absolute value such as 'real' or 'fake'. 

This modification further helps the generator to extract more detailed textures \cite{wang2018esrgan}. However, in order to function more like a standard discriminator, we need to use Relativistic average Discriminator (RaD) which represents the same concepts but as an average.

\subsection{Real-ESRGAN}
Real-Enhanced Super-Resolution Generative Adversarial Network is the extended version of ESRGAN which is trained with pure synthetic data \cite{wang2021realesrgan} and has practical uses in the restoration process as it better captures texture in real-world LR images. This causes training to become more challenging. 

The discriminator architecture utilizes U-NET structure and spectral normalization regularization. 

Real-ESRGAN presents two challenges in:
\begin{itemize}
    \item Training the network by covering unknown and complex degradation
    \item Creating one unified network
\end{itemize}

Real-ESRGAN synthesizes paired data with operations including blur, noise, and generalized trained models to real degradation. Indeed, it is a blind super-resolution with pure synthetic training pairs\cite{wang2021realesrgan}.  

\subsection{EDSR}

Enhanced Deep Super-Resolution network (EDSR) introduces a deep network architecture which is not involved in adversarial learning \cite{lim2017enhanced}. 
The model is an improved residual network which removes the BN layer to reach better performance by consuming less memory. This is achieved by stacking layers to increase performance which increases parameters and feature maps, causing the training procedure to become unstable \cite{lim2017enhanced}. 

To overcome this problem, a constant scaling factor is applied to each residual block after the second convolution layer. 

EDSR-BASE has a different structure compared to EDSR. Table 1 shows these differences. 
\begin{table}[h]
\centering
\caption{Compare the structure of EDSR and EDSR-BASE.}
\begin{tabular}{|p{1.2in}|p{0.7in}|p{0.8in}|}
\hline
\diagbox{config}{model}& EDSR & EDSR-BASE  \\ \hline
Data \underline{parallel}   & Yes & NO \\ \hline
Number of resblocks & 16 & 32 \\ \hline
Number of features  & 64 & 256 \\ \hline
\end{tabular}
\end{table}
\section{Methodology}
This section describes our approach to image degradation, valuation, and scoring techniques for GAN models. We used a pipeline for 

$\left[ Image Generation \to degradation \to SR-GAN optimization \to OCR \to scoring \right]$

\subsection{Character Recognition}
Optical Character Recognition (OCR) is a method to extract and recognize text out of images. In this work, we have used Tesseract, a leading open-source OCR engine that assumes the input is a binary image \cite{4376991}.

\subsection{Pipeline}
Our pipeline is designed to read text out of images. We have used different SR models to generate high quality images from low resolution ones, then use Tesseract for OCR and finally compare texts and score the performance of candidate models.

We follow steps shown in Fig. 1.

\begin{figure}[h]
    \centering
    \includegraphics[width=0.6\textwidth]{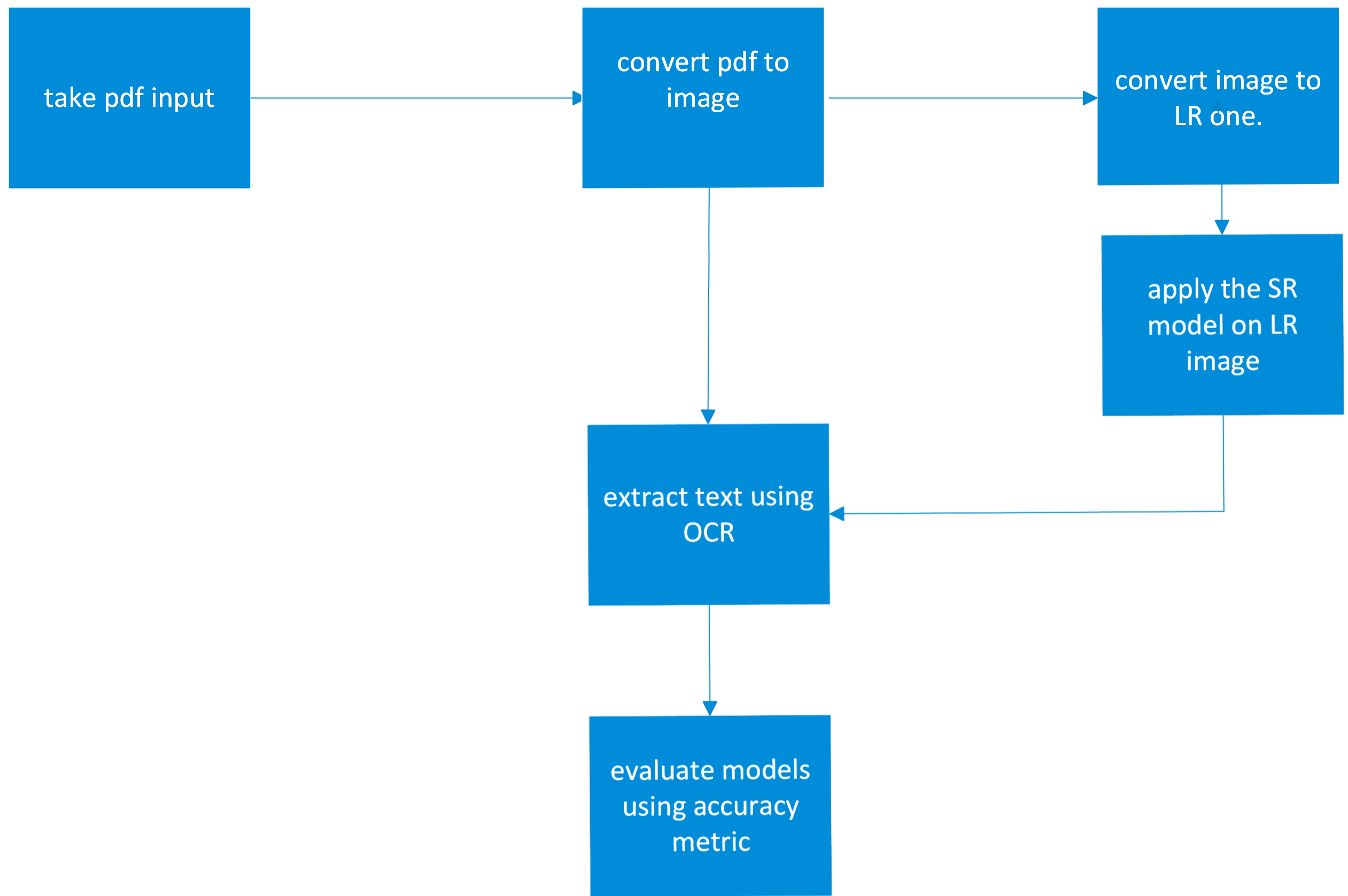}
    \caption{ Block diagram of pipeline.}
    \label{fig:mesh1}
\end{figure}

\section{Experimental Results}
We consider a text file covering different aspects of formatting including the font-size, font-style, font-weight (like bold, italics), text, etc. The image of the original text file including the low-resolution (LR) scales are provided below. We evaluate the capability of the candidate models in increasing the quality of the images. 

The results are compared based on text extracted (using Tesseract OCR). To assess the performance and accuracy in text comparison, we have used fuzz.ratio

\begin{figure}[htbp]
    \centering
    \includegraphics[width=1\textwidth]{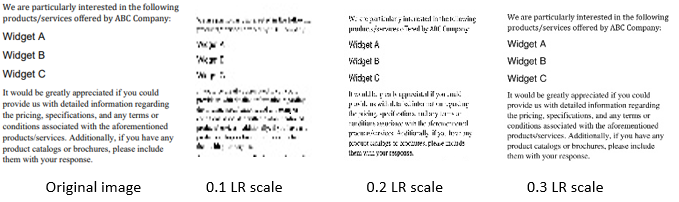}
    \label{fig:mesh2}
\end{figure}
\begin{figure}[htbp]
    \centering
    \includegraphics[width=1\textwidth]{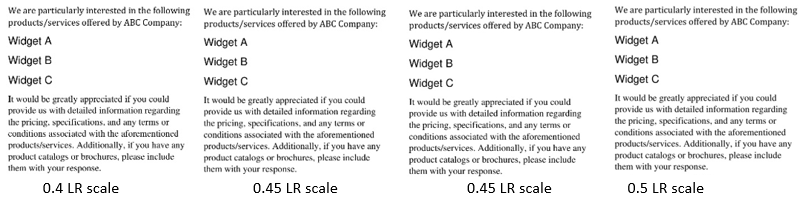}
    \caption{ Comparison of original 220 dpi image and degraded images with low-resolutions scales.}
    \label{fig:mesh3}
\end{figure}

Fuzzy string-matching\cite{fuzzywuzzy} uses Levenshtein distance (Levenshtein measures the minimum number of single-character edits needed to transform source string to target one) \cite{8054419}. 

We have used the fuzz ratio function from the fuzzywuzzy library to calculate accuracy. This speeds up the processing time between 4x and 10x. 

Tables 2 through 6 show accuracy of candidate models with images in 200 dpi, 220 dpi, 230 dpi, 240 dpi, 250 dpi and 260 dpi.

\begin{table}[htbp]
\centering
\caption{Accuracy of SR models after applying Tesseract on 200 dpi pdf image.}
\resizebox{\textwidth}{!}{%
\begin{tabular}{|p{1.2in}|p{1in}|p{1in}|p{0.8in}|p{1.2in}|}
\hline
\textbf{low resolution scale}  & \textbf{Hugging face EDSR-BASE }  &\textbf{Real-ESRGAN }   & \textbf{Esrgan}   &\textbf{Hugging face EDSR}   \\ \hline
0.1 &15\%   &25\%   &1\%   &12\%   \\ \hline
 0.2 &32\% &40\% &5\% &26\%   \\ \hline
 0.3 &85\% &87\% &67\% &83\%   \\ \hline
 0.35 &98\% &98\% &92\% &97\% \\ \hline
 \textbf{0.4} & \textbf{100\%} & 99\% & 98\% & 99\% \\ \hline
 \textbf{0.45} & \textbf{100\%} & \textbf{100\%} & 98\% & \textbf{100\%}   \\ \hline
 \textbf{0.5} & \textbf{100\%} & \textbf{100\%} & \textbf{100\%} & \textbf{100\%}   \\ \hline
\end{tabular}%
}
\end{table}

\begin{table}[htbp]
\centering
\caption{Accuracy of SR models after applying Tesseract on 220 dpi pdf image.}
\resizebox{\textwidth}{!}{%
\begin{tabular}{|p{1.2in}|p{1in}|p{1in}|p{0.8in}|p{1.2in}|}
\hline
\textbf{low resolution scale}  & \textbf{Hugging face EDSR-BASE }  &\textbf{Real-ESRGAN }   & \textbf{Esrgan}   &\textbf{Hugging face EDSR}   \\ \hline
0.1 &9\%   &27\%   &8\%   &10\%   \\ \hline
 0.2 &42\% &46\% &5\% &37\%   \\ \hline
 0.3 &92\% &92\% &86\% &90\%   \\ \hline
 0.35 &97\% &97\% &92\% &96\% \\ \hline
\textbf{ 0.4} & \textbf{100\%} & \textbf{100\%} & 98\% & \textbf{100\%} \\ \hline
 \textbf{0.45} & \textbf{100\%} & \textbf{100\%} & 98\% & \textbf{100\%}   \\ \hline
 \textbf{0.5} & \textbf{100\%} & \textbf{100\%} & \textbf{100\%} & \textbf{100\%}   \\ \hline
\end{tabular}%
}
\end{table}

\begin{table}[htbp]
\centering
\caption{Accuracy of SR models after applying Tesseract on 230 dpi pdf image.}
\resizebox{\textwidth}{!}{%
\begin{tabular}{|p{1.2in}|p{1in}|p{1in}|p{0.8in}|p{1.2in}|}
\hline
\textbf{low resolution scale}  & \textbf{Hugging face EDSR-BASE }  &\textbf{Real-ESRGAN }   & \textbf{Esrgan}   &\textbf{Hugging face EDSR}   \\ \hline
0.1 &15\%   &24\%   &0\%   &7\%   \\ \hline
 0.2 &42\% &48\% &5\% &39\%   \\ \hline
 0.3 &95\% &98\% &87\% &94\%   \\ \hline
 0.35 &99\% &99\% &96\% &98\% \\ \hline
 \textbf{0.4} & \textbf{100\%} & \textbf{100\%} & 98\% & \textbf{100\%} \\ \hline
 \textbf{0.45} & \textbf{100\%} & \textbf{100\%} & 98\% & \textbf{100\%}   \\ \hline
 \textbf{0.5} & \textbf{100\%} & \textbf{100\%} & 98\% & \textbf{100\%}   \\ \hline
\end{tabular}%
}
\end{table}

\begin{table}[htbp]
\centering
\caption{Accuracy of SR models after applying Tesseract on 240 dpi pdf image.}
\resizebox{\textwidth}{!}{%
\begin{tabular}{|p{1.2in}|p{1in}|p{1in}|p{0.8in}|p{1.2in}|}
\hline
\textbf{low resolution scale}  & \textbf{Hugging face EDSR-BASE }  &\textbf{Real-ESRGAN }   & \textbf{Esrgan}   &\textbf{Hugging face EDSR}   \\ \hline
0.1 &15\%   &29\%   &2\%   &19\%   \\ \hline
 0.2 &44\% &52\% &5\% &41\%   \\ \hline
 0.3 &94\% &98\% &89\% &93\%   \\ \hline
 0.35 &99\% &98\% &96\% &99\% \\ \hline
 \textbf{0.4} & \textbf{100\%} & \textbf{100\%} & 96\% & \textbf{100\%} \\ \hline
 \textbf{0.45} & \textbf{100\%} & \textbf{100\%} & 97\% & \textbf{100\%}   \\ \hline
 \textbf{0.5} & \textbf{100\%} & \textbf{100\%} & \textbf{100\%} & \textbf{100\%}   \\ \hline
\end{tabular}%
}
\end{table}
\begin{table}[htbp]
\centering
\caption{Accuracy of SR models after applying Tesseract on 250 dpi pdf image.}
\resizebox{\textwidth}{!}{%
\begin{tabular}{|p{1.2in}|p{1in}|p{1in}|p{0.8in}|p{1.2in}|}
\hline
\textbf{low resolution scale}  & \textbf{Hugging face EDSR-BASE }  &\textbf{Real-ESRGAN }   & \textbf{Esrgan}   &\textbf{Hugging face EDSR}   \\ \hline
0.1 &19\%   &28\%   &1\%   &12\%   \\ \hline
 0.2 &50\% &58\% &9\% &47\%   \\ \hline
 0.3 &96\% &97\% &93\% &96\%   \\ \hline
 0.35 &99\% &100\% &93\% &99\% \\ \hline
 \textbf{0.4} & \textbf{100\%} & \textbf{100\%} & 98\% & \textbf{100\%} \\ \hline
 \textbf{0.45} & \textbf{100\%} & \textbf{100\%} & 97\% & \textbf{100\%}  \\ \hline
 \textbf{0.5} & \textbf{100\%} & \textbf{100\%} & \textbf{100\%} & \textbf{100\%}   \\ \hline
\end{tabular}%
}
\end{table}
\begin{table}[htbp]
\centering
\caption{Accuracy of SR models after applying Tesseract on 260 dpi pdf image.}
\resizebox{\textwidth}{!}{%
\begin{tabular}{|p{1.2in}|p{1in}|p{1in}|p{0.8in}|p{1.2in}|}
\hline
\textbf{low resolution scale}  & \textbf{Hugging face EDSR-BASE }  &\textbf{Real-ESRGAN }   & \textbf{Esrgan}   &\textbf{Hugging face EDSR}   \\ \hline
0.1 &21\%   &30\%   &0\%   &19\%   \\ \hline
 0.2 &51\% &62\% &16\% &53\%   \\ \hline
 0.3 &99\% &99\% &94\% &98\%   \\ \hline
 \textbf{0.35} & \textbf{100\%} & \textbf{100\%} & 98\% & 99\% \\ \hline
 \textbf{0.4} & \textbf{100\%} & \textbf{100\%} & 96\% & \textbf{100\%} \\ \hline
 \textbf{0.45} & \textbf{100\%} & \textbf{100\%} & 97\% & \textbf{100\%}   \\ \hline
 \textbf{0.5} & \textbf{100\%} & \textbf{100\%} & \textbf{100\%} & \textbf{100\%}   \\ \hline
\end{tabular}%
}
\end{table}

Using the pipeline, we have tested all the candidate models and saw the following results. In the context of degrading an image to low-resolution (LR), we have applied six scales to reduce the quality to evaluate the model at each stage [indicative of noise and quality seen in images for OCR], with DPI resolution ranging from 200-260 dpi. Results are shown in the below tables. Images with dpi above 260 are clear enough for OCR to perform accurately. These results are not included in the tables below.

In Fig. 2, The images with LR scales of below 0.35 are hard to read by the human eye. This translates into poor OCR accuracy 

Tables 2 to 6 shows that scales lower than 0.3 severely degrades image quality and candidate models are unable to improve the quality sufficiently. 

\textbf{Real-ESRGAN} performs better than EDSR-BASE with scales of 0.1, 0.2 and 0.3, although these improvements are not able to attain 100\% OCR accuracy. 

\textbf{EDSR-BASE} is able to consistently generate higher accuracy than other candidate models with scales higher than 0.35. 

\section{Conclusion}
Our analysis was scored based on the accuracy seen in OCR text extraction using high-quality generated images across candidate models. Our analysis indicates that the best model meeting our requirement is \textbf{EDSR-BASE}. 

The choice between Real-ESRGAN and EDSR can vary depending on the specific requirement of the application, as Real-ESRGAN performs slightly better with more complex structure. 

EDSR-BASE outperforms Real-ESRGAN as it requires less computational power, which makes it more efficient when it comes to large-scale scenarios (as EDSR-BASE requires less compute to achieve 100\% OCR accuracy). 

EDSR-BASE gives us a consistent accuracy of 100\% for all the resolutions degraded at scales of 0.4.

EDSR-base compares to EDSR requires less compute resources. ESRGAN compared to Real-ESRGAN and EDSR was unable to remove blurs, complicated noise and compression artifacts which leads to decrease accuracy in OCR extraction on real-world images.

\section{Future areas of research}
This paper uses Tesseract as the OCR engine to compare the effectiveness of SR improvements from candidate models, however a different OCR engine might give different results for the same (scaled) inputs. 
This would be an area for future research.

\section{Acronyms}

SR GAN = Super Resolution Generative Adversarial Networks
LR = Low Resolution

RRDB = Residual-in-residual Dense Block

OCR = Optical Character Recognition

SR = Super Resolution

PReLU = Parametric Rectifier Linear Unit

BN = Bacth Normalization

RaD = Relativistic average Discriminator

\bibliographystyle{unsrt}
\bibliography{references}

\begin{thebibliography}{10}

\bibitem{tajik2021ideal}
Kimia Tajik.
\newblock Ideal thumbnail-preserving encryption for balancing image privacy and usability.
\newblock 2021.

\bibitem{lim2017enhanced}
Bee Lim, Sanghyun Son, Heewon Kim, Seungjun Nah, and Kyoung~Mu Lee.
\newblock Enhanced deep residual networks for single image super-resolution, 2017.

\bibitem{NIPS2014_5ca3e9b1}
Ian Goodfellow, Jean Pouget-Abadie, Mehdi Mirza, Bing Xu, David Warde-Farley, Sherjil Ozair, Aaron Courville, and Yoshua Bengio.
\newblock Generative adversarial nets.
\newblock In Z.~Ghahramani, M.~Welling, C.~Cortes, N.~Lawrence, and K.Q. Weinberger, editors, {\em Advances in Neural Information Processing Systems}, volume~27. Curran Associates, Inc., 2014.

\bibitem{goodfellow2014generative}
Ian~J. Goodfellow, Jean Pouget-Abadie, Mehdi Mirza, Bing Xu, David Warde-Farley, Sherjil Ozair, Aaron Courville, and Yoshua Bengio.
\newblock Generative adversarial networks, 2014.

\bibitem{eugenesiowEDSR}
Eugene Siow.
\newblock Edsr.
\newblock \url{https://huggingface.co/eugenesiow/edsr}, Year Accessed.

\bibitem{eugenesiowEDSRBase}
Eugene Siow.
\newblock Edsr-base.
\newblock \url{https://huggingface.co/eugenesiow/edsr-base}, Year Accessed.

\bibitem{wang2018esrgan}
Xintao Wang, Ke~Yu, Shixiang Wu, Jinjin Gu, Yihao Liu, Chao Dong, Yu~Qiao, and Chen~Change Loy.
\newblock Esrgan: Enhanced super-resolution generative adversarial networks.
\newblock In {\em The European Conference on Computer Vision Workshops (ECCVW)}, September 2018.

\bibitem{wang2021realesrgan}
Xintao Wang, Liangbin Xie, Chao Dong, and Ying Shan.
\newblock Real-esrgan: Training real-world blind super-resolution with pure synthetic data.
\newblock In {\em International Conference on Computer Vision Workshops (ICCVW)}, 2021.

\bibitem{ledig2017photorealistic}
Christian Ledig, Lucas Theis, Ferenc Huszar, Jose Caballero, Andrew Cunningham, Alejandro Acosta, Andrew Aitken, Alykhan Tejani, Johannes Totz, Zehan Wang, and Wenzhe Shi.
\newblock Photo-realistic single image super-resolution using a generative adversarial network, 2017.

\bibitem{ioffe2015batch}
Sergey Ioffe and Christian Szegedy.
\newblock Batch normalization: Accelerating deep network training by reducing internal covariate shift, 2015.

\bibitem{zhang2018image}
Yulun Zhang, Kunpeng Li, Kai Li, Lichen Wang, Bineng Zhong, and Yun Fu.
\newblock Image super-resolution using very deep residual channel attention networks, 2018.

\bibitem{zhang2018residual}
Yulun Zhang, Yapeng Tian, Yu~Kong, Bineng Zhong, and Yun Fu.
\newblock Residual dense network for image super-resolution, 2018.

\bibitem{jolicoeurmartineau2018relativistic}
Alexia Jolicoeur-Martineau.
\newblock The relativistic discriminator: a key element missing from standard gan, 2018.

\bibitem{4376991}
Ray Smith.
\newblock An overview of the tesseract ocr engine.
\newblock In {\em Ninth International Conference on Document Analysis and Recognition (ICDAR 2007)}, volume~2, pages 629--633, 2007.

\bibitem{fuzzywuzzy}
{Open Source Contributors}.
\newblock fuzzywuzzy.
\newblock \url{https://pypi.org/project/fuzzywuzzy/}, Year Accessed.

\bibitem{8054419}
Shengnan Zhang, Yan Hu, and Guangrong Bian.
\newblock Research on string similarity algorithm based on levenshtein distance.
\newblock In {\em 2017 IEEE 2nd Advanced Information Technology, Electronic and Automation Control Conference (IAEAC)}, pages 2247--2251, 2017.

\end{thebibliography}

\end{document}